\documentclass[letterpaper, 10 pt, conference]{ieeeconf}  

\IEEEoverridecommandlockouts                              
\usepackage[utf8]{inputenc}
\usepackage[T1]{fontenc}
\usepackage{cite}
\usepackage{amsmath,amssymb,amsfonts}
\usepackage{algorithmic}
\usepackage{gensymb}
\usepackage{graphicx}
\usepackage{hyperref}
\usepackage{textcomp}
\usepackage{xcolor}
\usepackage{multirow}
\usepackage{makecell}

\title{\LARGE \bf
RAVES-Calib: Robust, Accurate and Versatile Extrinsic Self Calibration Using Optimal Geometric Features
}

\author{ Haoxin Zhang$^{1,2,3}$, Shuaixin Li$^{2,3,\dagger}$, Xiaozhou Zhu$^{2,3}$, Xiao Zhang$^{2,3,4}$, Hongbo Chen$^{1}$, Wen Yao$^{2,3,\ast}$
\thanks{This research was funded by the National Natural Science Foundation of China(Grant No.42201501)}
\thanks{$^1$ authors with the School of Systems Science and Engineering, Sun Yat-Sen University, Guangzhou, China}
\thanks{$^2$ authors with the Defense Innovation Institute, Chinese Academy of Military Science, Beijing, China}
\thanks{$^3$ authors with the Intelligent Game and Decision Laboratory, Beijing, China}
\thanks{$^4$ authors with the College of Computer Science and Technology, Harbin Engineering University, Harbin, China}
\thanks{$^\dagger$ authors contributed equally}
\thanks{$^\ast$ corresponding authors (e-mail: wendy0782@126.com)}
}

\begin{document}
\maketitle
\thispagestyle{empty}
\pagestyle{empty}

\begin{abstract}
In this paper, we present a user-friendly LiDAR-camera calibration toolkit that is compatible with various LiDAR and camera sensors and requires only a single pair of laser points and a camera image in targetless environments. Our approach eliminates the need for an initial transform and remains robust even with large positional and rotational LiDAR-camera extrinsic parameters. We employ the Gluestick pipeline to establish 2D-3D point and line feature correspondences for a robust and automatic initial guess. To enhance accuracy, we quantitatively analyze the impact of feature distribution on calibration results and adaptively weight the cost of each feature based on these metrics. As a result, extrinsic parameters are optimized by filtering out the adverse effects of inferior features. We validated our method through extensive experiments across various LiDAR-camera sensors in both indoor and outdoor settings. The results demonstrate that our method provides superior robustness and accuracy compared to SOTA techniques. Our code is open-sourced on GitHub\footnote{\url{https://github.com/haoxinnihao/LiDAR_camera_calibration}} to benefit the community.


\end{abstract}

\section{INTRODUCTION}


With the advancement of autonomous systems, intelligent agents increasingly require detailed environmental information. LiDAR and cameras are two essential sensors that provide complementary data, and they are widely employed in various autonomous system applications, e.g., mobile mapping\cite{puente2013review}, localization\cite{lin2024r}, obstacle detection\cite{4432828} and environmental perception\cite{10089400}. While cameras capture high-resolution images rich in color and texture details, they are highly sensitive to lighting conditions\cite{yang2015automatic}. In contrast, LiDAR measurements remain unaffected by lighting variations and provide accurate distance information. However, LiDAR suffers from a narrower field of view (FoV) and lower resolution compared to cameras\cite{li2019nrli}. To fully exploit the strengths of both sensors, robust, accurate, and convenient calibration of their extrinsic parameters is critical for effective sensor fusion, which can unlock their combined potential.

\begin{figure}[t]
\centerline{\includegraphics[width=\linewidth]{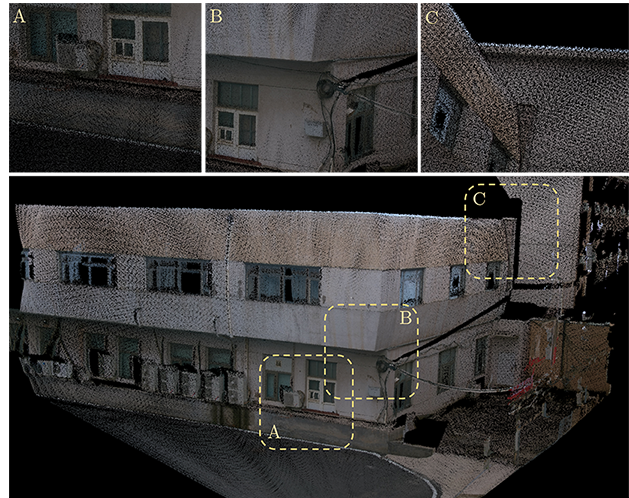}}
\caption{Colorized LiDAR points by camera RGB image with calibrated extrinsic parameters using the proposed method. The second row is the overall color point cloud, and the first row is the enlarged picture of the local details in the color point cloud. }
\label{fig1}
\end{figure}

Existing high-precision LiDAR-camera calibration methods primarily rely on external targets, such as checkerboards or specific patterns, and involve laborious manual processes\cite{huang2024novel,domhof2021joint,yi2022extrinsic,liao2018extrinsic}. These cooperative targets, with well-defined patterns, shapes, and dimensions, facilitate the extraction of precise mutual features from the one-shot of LiDAR point cloud and camera image. These features are then used to estimate the extrinsic parameters by minimizing the reprojection error. Conversely, targetless calibration methods rely on naturally occurring features, making their performance highly dependent on the quality of the feature extraction and matching. 

We argue that a user-friendly calibration toolkit must prioritize two key aspects: convenience and accuracy. For convenience, the calibration algorithm should function across various environments without requiring specific targets, should be highly automated, and should support different sensor types. In terms of accuracy, the results must match or exceed the precision achieved by target-based methods while maintaining the benefits of automation and flexibility.

Despite advancements, significant challenges remain. In LiDAR-camera calibration, accurate heterogeneous data association is fundamental for solving extrinsic parameters, and its quality directly affects the calibration results. However, the complexity of natural environments, sensor data heterogeneity, and parallax effects limit the accuracy of feature extraction and matching. Additionally, the location of features in natural settings is random, which can lead to inferior distribution of extracted features adversely affecting the accuracy and stability of extrinsic parameter estimation.

To address the aforementioned challenges, we propose RAVES-Calib, a robust, accurate and versatile self-calibration toolkit for fully automatic LiDAR-camera extrinsic parameter calibration in both indoor and outdoor environments, without the need for artificial targets. The toolkit is developed based on the outstanding work in \cite{yuan2021pixel} and enhanced by leveraging the state-of-the-art (SOTA) deep learning model, Gluestick, to extract and match stable geometric features providing a rough initial estimate and eliminating dependence on initial values. Furthermore, we analyze the impact of feature distribution on calibration accuracy and use this information to guide optimal feature selection. Both point and line features are incorporated to compensate for regions with insufficient features, ensuring reliable calibration across diverse conditions. Our main contributions are as follows:

\begin{itemize}
\item A novel heuristic-free optimal features selection algorithm is developed to reliably screen and select a subset of geometric features with even distribution among all matching 2D-3D correspondences.
\item A systematic targetless LiDAR-camera extrinsic calibration pipeline is designed on the basis of the filtering of the optimal group of point and line features for a single pair of point cloud and image. 
\item An open-sourced LiDAR-camera calibration toolkit is developed which does not require an initial transform and is compatible with various sensor types. Extensive experiments conducted in both indoor and outdoor environments demonstrate that the proposed toolkit is highly consistent across different calibration scenarios, achieving accuracy on par with, or better than, target-based methods in natural environments.
\end{itemize}

\section{Related Work}
LIDAR-camera extrinsic calibration has been extensively studied in the field of autonomous systems and can be simply categorized into two categories: target-based and targetless. 

\subsection{Target-based Method}
Target-based methods represent the classic solution to solving the LiDAR-camera extrinsic calibration problem, particularly due to the challenge to associating features from heterogeneous data. These methods leverage well-defined patterns and shapes, which allow for the precise establishment of geometric constraints, such as plane normals and chessboard edges. The main task of these methods is the high-precision feature extraction and matching of features. For instance, in \cite{park2014calibration,kwak2011extrinsic,liao2018extrinsic}, polygons are used as calibration targets. Both the camera and LiDAR extract corner points or edges from these targets and then perform 2D-3D data association to estimate extrinsic parameters by minimizing reprojection error. In \cite{6232233}, a cardboard with an ArUco marker is used, enabling both the camera and LiDAR to directly extract the 3D positions of the target's corner points for 3D-3D data association. In \cite{mirzaei20123d} and \cite{cui2023aclc}, checkerboards serve as calibration targets. \cite{mirzaei20123d} estimates extrinsic parameters by aligning normal vectors of the target between LiDAR and camera data, while leverages the LiDAR intensity image to establish 2D-2D data associations, followed by back-projecting the checkerboard corner points from the intensity map into the point cloud for final data association. \cite{huang2024novel} uses a painted acrylic checkerboard target, allowing laser beams to pass through the “white” grid areas and direct matching detected grids in image and point cloud.

While target-based methods can achieve high precision through evenly distributed and accurately matched features, they heavily depend on calibration targets, which are often cumbersome to manufacture and transport. Furthermore, these methods do not satisfy the requirements for automated calibration processes.

\subsection{Targetless method}
Targetless method directly extracts stable features from the environment, such as lines, planes and semantic objects. For example, in \cite{yuan2021pixel}, \cite{wu2022lidar}, \cite{bai2020lidar}, \cite{munoz2020targetless}, edge features from LiDAR point clouds are projected onto the corresponding edge features in camera RGB images, and the feature distance or point-to-line distance is minimized to estimate extrinsic parameters. \cite{koide2023general} utilizes a deep learning method to extract matching feature points between the LiDAR intensity image and the camera’s RGB image, uses reprojection and RANSAC for initial value estimation, and further refines the results using the minimum posterior mutual information distance \cite{mastin2009automatic}. In \cite{huang2024onlinetargetfreelidarcameraextrinsiccalibration}, researchers utilize the robustness of semantic segmentation to extract features from both the LiDAR intensity and the camera RGB images, selecting corner points of each semantic region as calibration features.

However, acquiring high-precision, accurately matched, and evenly distributed features from the environment remains a significant challenge due to environmental variability. As a result, targetless methods often struggle to match the accuracy achieved by target-based methods\cite{li2023automatic}. \cite{yuan2021pixel} uses the method of traversal within a certain range to determine the initial estimate extrinsic parameter, which is not only time-consuming, but also cannot obtain the usable result when the extrinsic parameter is outside the traversal range. \cite{koide2023general} obtains a robust initial estimate by using deep learning features, the accuracy of the refining results using normalized information distance needs to be improved. We improved the initial estimation performance under the framework of \cite{koide2023general}, refined with multiple types of features, and analyzed the sources of calibration errors. Specifically, the distribution of feature points plays a critical role in calibration stability and accuracy\cite{SEO2004733,X-ICP}, yet this aspect remains underexplored in existing calibration research.


\section{METHODOLOGY}
\begin{figure*}[!t]
\includegraphics[width=\linewidth]{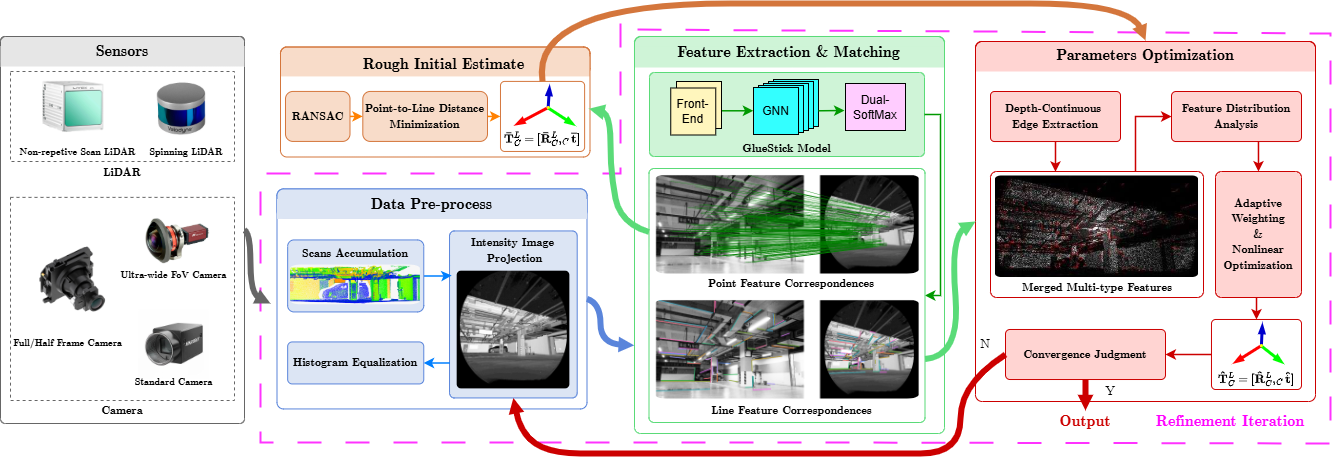}
\caption{The pipeline of LiDAR-camera calibration process.}
\label{fig2}
\end{figure*}

\subsection{Overview}\label{AA}
Fig.\ref{fig2} illustrates the pipeline of the proposed RAVES-Calib toolkit. Let $L$ and $C$ denote the frames of the LiDAR and camera, respectively. The extrinsic calibration problem is formulated as estimating the transformation $\mathbf{T}_C^L = (\mathbf{R}_C^L, _C\mathbf{t}), \mathbf{R}_C^L \in \mathrm{SO(3)},  _C\mathbf{t} \in \mathbb{R}^3$ from $C$ to $L$ given a pair of laser points $\mathcal{P}$ and a camera image $\mathcal{I}$ sampled simultaneously. The proposed calibration toolkit comprises four main modules: rough initial estimate, data pre-processing, feature extraction and matching, and extrinsic parameters refinement. In the initial phase, data are denoised and rectified before being passed into the Gluestick algorithm \cite{pautrat2023gluestick} to establish 2D-3D feature correspondences and estimate an initial transformation $\mathbf{\bar{T}}_C^L$. The distribution of matched correspondences is quantitatively analyzed, enabling adaptive adjustment of each correspondence's influence on the optimization process. Finally, the accurate extrinsic parameters are refined through iterative optimization until convergence.

\subsection{Rough Initial Estimate} 

For non-repetitive LiDARs, the point cloud $\mathcal{P}$ is accumulated statically, whereas for spinning LiDARs, it is necessary to move LiDARs in a small area and then the point cloud $\mathcal{P}$ is constructed using laser odometry (we use continuous-time ICP) to build a local map. The intensity image $\mathcal{I}_\mathcal{P}$ is generated by projecting the points $_L\mathbf{P} \in \mathbb{R}^3$ in $\mathcal{P}$ onto the pixel coordinates of $\mathcal{I}$, where the pixel intensity corresponds to the point's intensity value. A mapping relationship $\Upsilon = \mathbf{p}_i \in \mathcal{I} \mapsto \mathbf{P}_k \in \mathcal{P}, k = \left | \mathbf{p}_i \right | $ between points $_L\mathbf{P}$ in $\mathcal{P}$ and pixel $\mathbf{p}$ in $\mathcal{I}_\mathcal{P}$ is created. This means that for a given pixel $\mathbf{p}_i$, a corresponding set of points $_L\mathbf{P}{k}, k \in \left | \mathbf{p}_i \right |$ can be indexed.

\begin{figure}[t]
\centerline{\includegraphics[width=\linewidth]{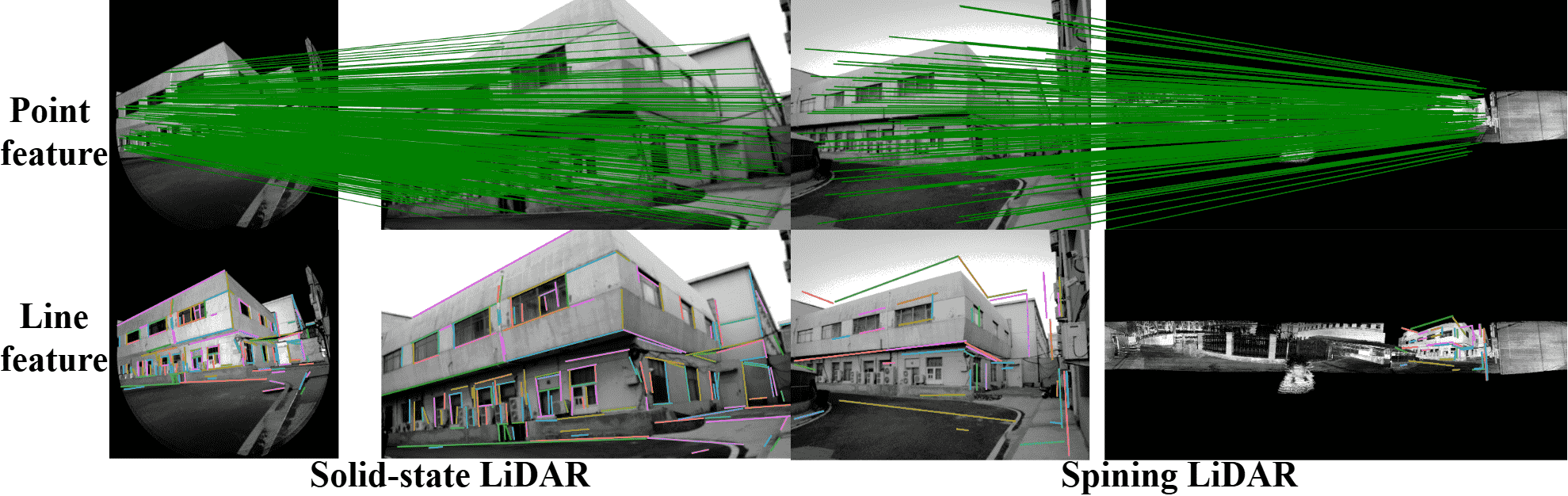}}
\caption{Gluestick can find the correspondence between point and line features in LiDAR intensity image and camera images.}
\label{fig3}
\end{figure}

Gluestick can extract and match point and line features in two images, as shown in Fig.\ref{fig3}. We use Gluestick to establish feature correspondences between $\mathcal{I}$ and $\mathcal{P}$. Initially, only point features are utilized to estimate the rotational component $\mathbf{\bar{R}}_C^L$ of $\mathbf{T}_C^L$ using random Random Sample Consensus (RANSAC) and Perspective-n-Point (PnP) algorithms to manage outlier correspondences. Then line features are incorporated to formulate a nonlinear optimization problem, which is solved by minimizing the point-to-line distance. Specifically, let $_L\mathbf{P}_{i} \in \mathbb{R}^3$ denote a 3D line feature point in $\mathcal{P}$, obtained according to $\Upsilon(\cdot)$. Meanwhile, $\mathbf{n}_{j} \in \mathbb{R}^2$ and $\mathbf{q}_{j} \in \mathbb{R}^2$ denote the 2D normal and a point on the corresponding line feature in $\mathcal{I}$. Ideally, the projection of $_L\mathbf{P}_i$ onto the image plane using the estimated rotation $\mathbf{\bar{R}}_C^L$ should align with the edge feature $(\mathbf{n}_j, \mathbf{q}_j)$ in $\mathcal{I}$. The equation is formulated as follows:
\begin{equation}
0 = \overbrace{\mathbf{n}^\top_j \cdot ( \underbrace{\mathbf{f} ( \boldsymbol{\pi} (\underbrace{\mathbf{T}_C^L \cdot _L\mathbf{P}_i }_{_C\mathbf{P}_i } ))}_{\mathbf{p}_i=[u_i, v_i]^\top } -  \mathbf{q}_{j})}^{\text{point-to-line distance}}\label{eq1}
\end{equation}
where $\mathbf{f}(\cdot)$ denotes the camera distortion model, $\boldsymbol{\pi}(\cdot)$ denotes the camera projection model.

\subsection{Feature Distribution Analysis}
\subsubsection{Calibration Optimization Establishment}
To compensate for regions with insufficient features density, we use point and line features concurrently. For point features, the cost is represented by the standard reprojection error $\mathbf{e}^p$. For line features, the point-to-line distance described in e.q.(\ref{eq1}) is used as the cost term $\mathbf{e}^l$. Note that, the midpoint of the edge feature segment is chosen as the point $\mathbf{q}_j$ lying on the line feature in $\mathcal{I}$. The least squares problem for calibration can thus be formulated as follows: 
\begin{equation}
\mathop{\min}_{\mathbf{T}^{L}_{C}}{\Vert \mathbf{e}^p  \Vert}_{2} + {\Vert \mathbf{e}^l  \Vert}_{2}  \label{eq2}
\end{equation}

\subsubsection{Feature distribution analyze}
As mentioned in \cite{X-ICP}, the Hessian matrix can reflect the contribution of features to the ICP algorithm for positioning. Similarly, the Hessian matrix can also reflect the contribution of the relative positioning between the radar and the camera. To explore the impact of feature distribution on calibration, we analyze the Hessian matrix $\mathbf{H} = \mathbf{J}^\top \mathbf{J}$ of the cost function, where $\mathbf{J} = [\mathbf{J}^p, \mathbf{J}^l]^{\top}$ is a $(2N + M) \times 6$ Jacobian matrix. Here, $N$ and $M$ represent the number of point and line features, respectively. For a specific point feature $i$ and line feature $j$, the Jacobian matrices are defined as:
\begin{equation}
\mathbf{J}_i^p = \bf{\Phi}(\mathrm{_L}\mathbf{P}_i), \quad \mathbf{J}_j^l = \mathbf{n}_j^\top \cdot \bf{\Phi}(\mathrm{_L}\mathbf{P}_j)
\end{equation}
where $\bf{\Phi}(\mathrm{_L}\mathbf{P})$ is the standard reprojection Jacobian evaluated at $\mathrm{_L}\mathbf{P} = [X,Y,Z]^{\top}$:
\begin{equation}
\bf{\Phi}(\mathrm{_L}\mathbf{P}) = \left[
\begin{matrix}
  -\frac{f_{x}XY}{Z^{2}} & -f_{y}-\frac{f_{y}Y^{2}}{Z^{2}}\\
  f_{x}+\frac{f_{x}X^{2}}{Z^{2}} & \frac{f_{y}X}{Z}\\
  -\frac{f_{x}Y}{Z} & 0 \\
  \frac{f_{x}}{Z} & \frac{f_{y}XY}{Z^{2}}\\
  0 & \frac{f_{y}}{Z}\\
  -\frac{f_{x}X^{2}}{Z^{\prime2}} & -\frac{f_{x}Y^{2}}{Z^{2}}
\end{matrix}
\right]^{\top}
\end{equation}
The Jacobian is composed of rotational and positional parts:
\begin{equation}
    \underset{2 \times 6}{\mathbf{J}^p} = [\underset{2 \times 3}{\mathbf{J}^p_r}, \underset{2 \times 3}{\mathbf{J}^p_t}],\quad \underset{1 \times 6}{\mathbf{J}^l} = [\underset{1 \times 3}{\mathbf{J}^l_r}, \underset{1 \times 3}{\mathbf{J}^l_t}]
\end{equation}
The corresponding $6 \times 6$ Hessians are denoted as ${\mathbf{H}_i^p}$ and $\mathbf{H}_j^l$. The complete Hessian can be represented as block matrices:

\begin{equation}
\mathbf{H} = \begin{bmatrix} \mathbf{H}_{rr} & \mathbf{H}_{rt} \\ \mathbf{H}_{tr} & \mathbf{H}_{tt} \end{bmatrix}
\end{equation}
Here, $\mathbf{H}_{rr} \in \mathbb{R}^{3\times3}$ contains information related to the rotational component, while $\mathbf{H}_{tt} \in \mathbb{R}^{3\times3}$ relates to the translational component. The coupling between the rotation $\mathbf{R} \mapsto \mathbf{r} \in \mathfrak{so}(3)$ and translation $\mathbf{t} \in \mathbb{R}^3$ is nontrivial since differences in scale and type\cite{X-ICP}. Thus, only $\mathbf{H}_{rr}$ and $\mathbf{H}_{tt}$ are subjected to SVD decomposition for eigen-analysis:
\begin{equation}
 \mathbf{H}_{tt} = \mathbf{V}_{t}\boldsymbol{\Sigma}_{t}\mathbf{V}^{\top}_t, \quad \mathbf{H}_{rr} = \mathbf{V}_{r}\boldsymbol{\Sigma}_{r}\mathbf{V}^{\top}_r \label{eq}
\end{equation}
where $\mathbf{V}_{t}$ and $\mathbf{V}_{r}$ are the eigenvector matrices, and $\mathbf{\Sigma}_{t}$ and $\mathbf{\Sigma}_{r}$ are diagonal matrices with the corresponding eigenvalues as the diagonal entries. As discussed in \cite{1240258}, while the eigenvalues in $\mathbf{\Sigma}_{r}$ and $\mathbf{\Sigma}_t$ provide a direct measure of the information along each eigenvector, these eigenvalues may not behave consistently across different environments and sensors, and thus cannot be directly indicate feature distribution. A formal relationship between features and optimization cost is necessary to evaluate their contribution. In degraded scene localization, the elements of the Jacobian matrix are used as a measure of the contribution to the cost \cite{X-ICP},\cite{10.1115/1.4031335},\cite{8301593}, given the established co-relationship between the Jacobian matrix and eigenvalues. This approach simplifies the analysis while maintaining relevance to the optimization Jacobian across varying environments.

\begin{figure}[t]
\centerline{\includegraphics[width=0.8\linewidth]{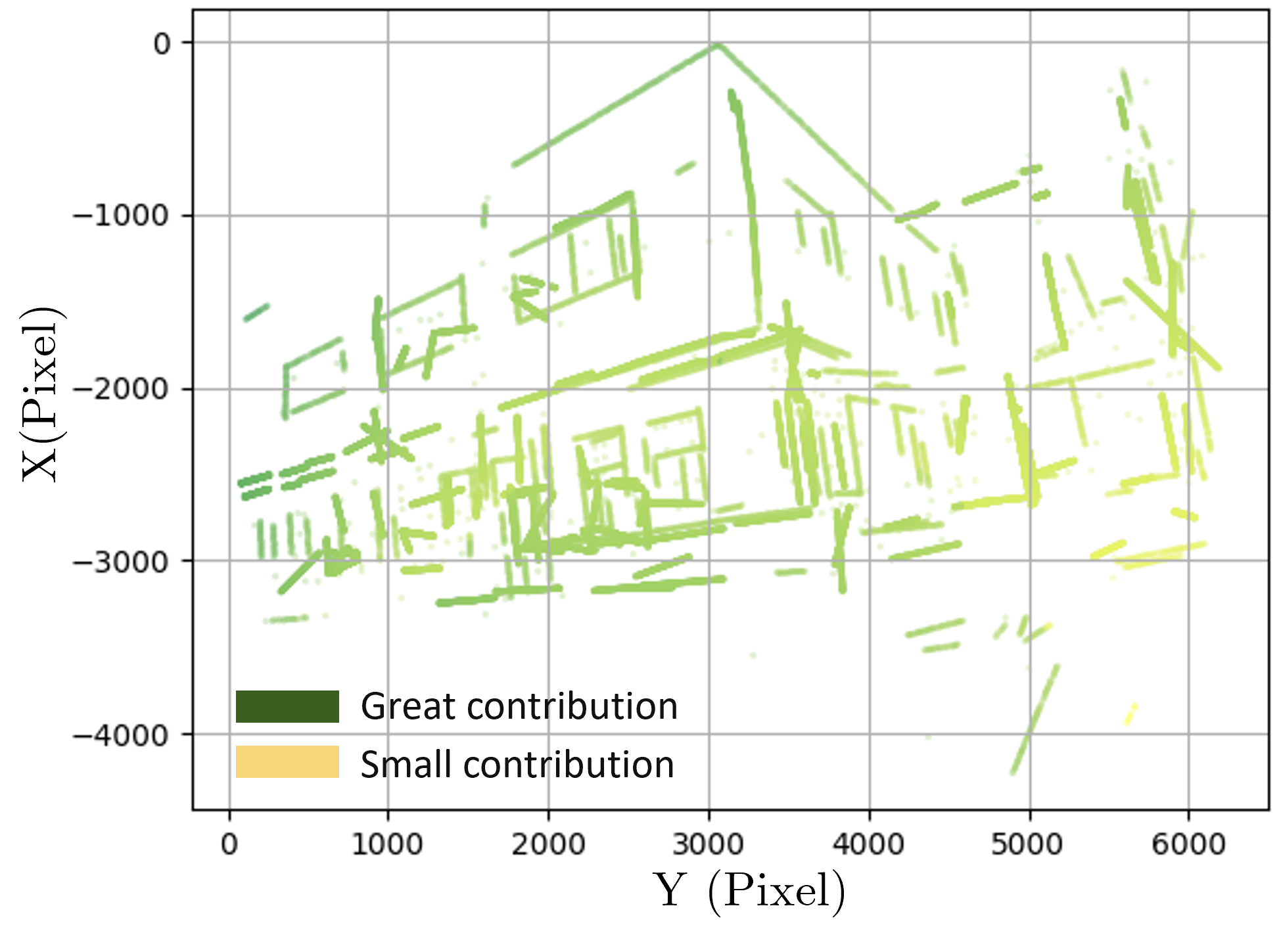}}
\caption{Feature contribution heatmap. The horizontal and vertical axes represent the x and y axes of the image, respectively, in pixels. Green represents high contribution, yellow represents low contribution.}
\label{fig4}
\end{figure}

Inspired by this concept, we propose a new feature distribution index to quantify each feature's contribution to the LiDAR-camera calibration result. The Jacobian matrix is normalized separately for the translation and rotation components. The L2 norm is utilized for the vector $\mathbf{J}_r^l$ and matrix $\mathbf{J}_r^p$, as this approach is more robust against scale differences in normalization \cite{10.1115/1.4031335}. The normalization process is defined as follows:
\begin{equation}
\mathcal{F}_{\mathcal{S}}^{\mathcal{T}}= \left[ \frac{\mathbf{J}_{\mathcal{S},1}^{\mathcal{T}}}{\Vert \mathbf{J}_{\mathcal{S},1}^{\mathcal{T}} \Vert_2} \cdots \frac{\mathbf{J}_{\mathcal{S},n}^{\mathcal{T}}}{\Vert \mathbf{J}_{\mathcal{S},n}^{\mathcal{T}} \Vert_2} \right]^{\top}\\
\label{eq6}
\end{equation}

To evaluate feature contribution along each direction in the six degrees of freedom (6DOF) space, the contributions defined in e.q.(\ref{eq6}) are projected onto the directions of the eigenvectors as follows:
\begin{equation}
\mathcal{D}_{\mathcal{S}}^\mathcal{T} = \left( \mathcal{F}_{\mathcal{S}}^\mathcal{T} \cdot \mathbf{V}_{\mathcal{S}}^\mathcal{T} \right)^{|\cdot |} \\ \label{eq7}
\end{equation}
In e.q.(\ref{eq6})-(\ref{eq7}), let $\mathcal{T} \in {l, p}$, and $\mathcal{S} \in {r, t}$. Here, $\mathcal{D}_{r}^l$, $\mathcal{D}_{t}^l$, $\mathcal{D}_{r}^p$,$\mathcal{D}_{t}^p$ are calibration contributions for all features, projected by the eigenvectors in ${\mathbf{V}_{r}, \mathbf{V}_{t}}$. The ${|\cdot |}$ operator indicates the elementwise absolute value of the vector. Concurrently, the scalar values in $\mathcal{D}_{r}^l$ and $\mathcal{D}_{t}^l$, as well as L1 norms of $\mathcal{D}_{r}^p$ and $\mathcal{D}_{t}^p$, are direct indicators of calibration contribution of a certain direction. Feature contributions are visualized with a heat map, as shown in Fig.\ref{fig4}.

\subsection{Optimal Feature Group-based Refinement}

\begin{figure}[t]
\centerline{\includegraphics[width=0.85\linewidth]{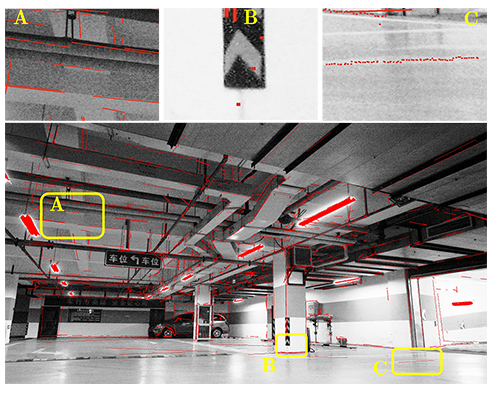}}
\caption{Multi-type feature projection image. A, Depth-continuous edge features. B, Intensity image point features. C, Intensity image line feature.}
\label{fig5}
\end{figure}

\subsubsection{Multi-type Feature Extraction}
Using a single type of feature is often insufficient for diverse calibration scenarios. As demonstrated in \cite{yuan2021pixel}, relying on a single feature type can lead to uneven feature distribution and a single edge feature orientation, complicating the calibration process. To address this issue, our approach, illustrated in Fig.\ref{fig5}, leverages multiple types of features: point features, line features corresponding to intensity images, and Depth-continuous edge features. This combination increases the robustness and adaptability of the calibration system across different environments. The correspondence between feature points is obtained through Gluestick, the corresponding Depth-continuous edge features and line features are obtained by KD-tree.

\begin{itemize}

\item \textbf{Depth-continuous edge feature}: Depth-continuous edges extract 3D structures directly from the point cloud, minimizing the impact of parallax and emphasizing the structured edges of objects in the environment. Although curved features can be obtained with appropriate voxel size selection, it is difficult to accurately capture a circular object because it cannot fit two intersecting planes, and in scenes dominated by unidirectional features, it may fail to provide sufficient constraints to determine the extrinsic parameters reliably. 

\item \textbf{Line features from Intensity images}: Intensity images of laser points offer texture information similar to that of RGB images. By applying Gluestick network, we extract line features that correspond to both intensity and RGB images. These line features capture variations in the material properties of objects in the scene. As a result, they complement Depth-continuous edge features by accurately capturing edges of circular objects and providing more detailed feature information.

\item \textbf{Intensity image point features}: In environments where only edge features in a single direction are available, each line feature provides constraints on only four degrees of freedom, limiting the calibration accuracy. To address this limitation, we introduce point features, which increase the number of constraints, particularly in situations where feature directions are uniform. This enhancement improves the robustness and reliability of the calibration process.
\end{itemize}

\subsubsection{Adaptive Weighting Scheme}
As previously discussed, we evaluate the contribution of each feature to the 6DOF in the calibration system. The total contribution of a feature is determined by summing its contributions across all directions. This total contribution is then incorporated as a weight into the cost function, reflecting the importance of each feature in the calibration process. Specifically, if the scalar value of a feature's contribution approaches $0$, it indicates that the matching pair provides little to no useful information for calibration. In such cases, the feature is down-weighted by assigning a near-zero value, effectively excluding it from the optimization process.
\begin{equation}
    \begin{aligned}
    \hat{\boldsymbol{\xi}} & = \mathop{\mathrm{argmin}}_{\hat{\boldsymbol{\xi}}}({\Vert \mathbf{e}^p  \Vert}_{\mathcal{W}^p} + {\Vert \mathbf{e}^l  \Vert}_{\mathcal{W}^l})
    \\ & = (\mathrm{J}^{\top}\mathcal{W}\mathrm{J})^{-1}\mathrm{J}^{\top}\mathcal{W}\mathbf{e} 
    \end{aligned}
\label{eq7}
\end{equation}

where $\boldsymbol{\xi}$ is the Lie algebra $\mathfrak{se}(3)$ of the transformation $\mathbf{T}_C^L$. The element of $\mathcal{W}$ is the metric introduced in e.q.(\ref{eq7}).

\section{EXPERIMENT}
\begin{figure}[tb]
\centerline{\includegraphics[width=0.45\linewidth]{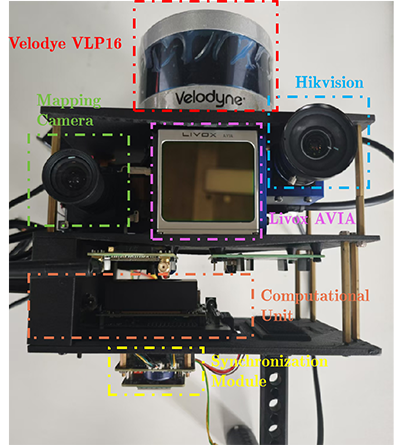}}
\caption{Our sensor suite, consisting of two cameras (SHARE and Hikvision) and two LiDARs (Livox and Velodyne), powered by Jetson Xavier NX. Note that the resolution of the SHARE full-frame camera is $6144\times4168$.}
\label{fig6}
\end{figure}

To validate RAVES-Calib, we conducted extensive experiments using real-world data collected from both indoor and outdoor scenarios. Our data collection platform (see Fig.\ref{fig6}) includes a Livox Avia solid-state LiDAR, a Velodyne VLP-16 mechanical spinning LiDAR, a SHARE full-frame camera, and a Hikvision industrial pinhole camera. Full-frame cameras have a larger sensor size and have higher resolution and image size. Tab.\ref{tab1} provides detailed sensor parameters, including models and fields of view (FoV). we recorded 8 pairs of LiDAR points and camera images in various environments, and ran the proposed calibration toolkit to evaluate results. Note that the Gluestick model used for feature extraction in the system is pre-trained.

\begin{table}[bt]
\caption{Detailed parameters of our data collection platform.}
\begin{center}
\resizebox{8cm}{!}{
\begin{tabular}{|c|c|c|c|}
\hline
\textbf{Sensor Model} & \textbf{Sensor Type}&\multicolumn{2}{|c|}{\textbf{FoV(\degree)}} \\
\cline{3-4} 
\textbf{} & \textbf{\textit{}}& \textbf{\textit{H}}& \textbf{\textit{V}} \\
\hline
Solid-state LiDAR& Livox AVIA & 70.4 & 77.2 \\
Spinning LiDAR& Velodyne VLP16 & 360 & 30 \\
Full Frame Camera& SHARE S2600 & 73 &  52\\
Pinhole Cameras& Hikvision CS016& 41.7 & 28.6 \\
\hline

\end{tabular}
}

\label{tab1}
\end{center}
\end{table}

\begin{table}[bt]
\caption{Detailed parameters of our data collection platform.}
\begin{center}
\resizebox{8.6cm}{!}{
\begin{tabular}{|c|c|c|c|c|c|c|c|}
\hline
\textbf{sensor}&\textbf{Method} &\textbf{01} &\textbf{02} &\textbf{03} &\textbf{04} &\textbf{05} &\textbf{06}  \\

\hline
\multirow{3}*{\text{\makecell{Solid-state\\Full Frame}}}&RAVES&$\checkmark$& $\checkmark$ & $\checkmark$ & $\checkmark$&$\checkmark$&$\checkmark$ \\
~&Ref\cite{koide2023general}&$\checkmark$& $\checkmark$& $\checkmark$ & $\checkmark$&$\checkmark$ & $\checkmark$\\
~&Ref\cite{yuan2021pixel}&$\times$& $\times$ & $\checkmark$ &  $\checkmark$ & $\times$& $\times$\\
\multirow{3}*{\text{\makecell{Solid-state \\Pinhole Camera }}}&RAVES&$\checkmark$& $\checkmark$ & $\checkmark$ & $\checkmark$& $\checkmark$& $\checkmark$\\
~&Ref\cite{koide2023general}&$\checkmark$& $\checkmark$ & $\checkmark$ & $\checkmark$& $\checkmark$& $\times$\\
~&Ref\cite{yuan2021pixel}&$\times$& $\times$ & $\checkmark$ & $\checkmark$& $\times$& $\times$ \\
\multirow{3}*{\text{\makecell{Spinnin \\Full Frame }}}&RAVES&$\checkmark$& $\checkmark$ & $\checkmark$ & $\checkmark$& $\checkmark$& $\times$\\
~&Ref\cite{koide2023general}&$\checkmark$& $\checkmark$ & $\checkmark$ & $\checkmark$& $\checkmark$& $\times$\\
~&Ref\cite{yuan2021pixel}&$\times$& $\times$ & $\checkmark$ & $\checkmark$& $\times$& $\times$ \\
\multirow{3}*{\text{\makecell{Spinnine \\Pinhole Camera}}}&RAVES&$\checkmark$& $\checkmark$ & $\checkmark$ & $\checkmark$& $\times$& $\times$ \\
~&Ref\cite{koide2023general}&$\checkmark$& $\checkmark$ & $\checkmark$ & $\checkmark$& $\times$& $\times$\\
~&Ref\cite{yuan2021pixel}&$\times$ & $\times$ & $\checkmark$ & $\checkmark$ & $\times$& $\times$\\
\hline

\end{tabular}
}

\label{tab2}
\end{center}
\end{table}


\begin{figure}[htbp]
\centerline{\includegraphics[width=0.95\linewidth]{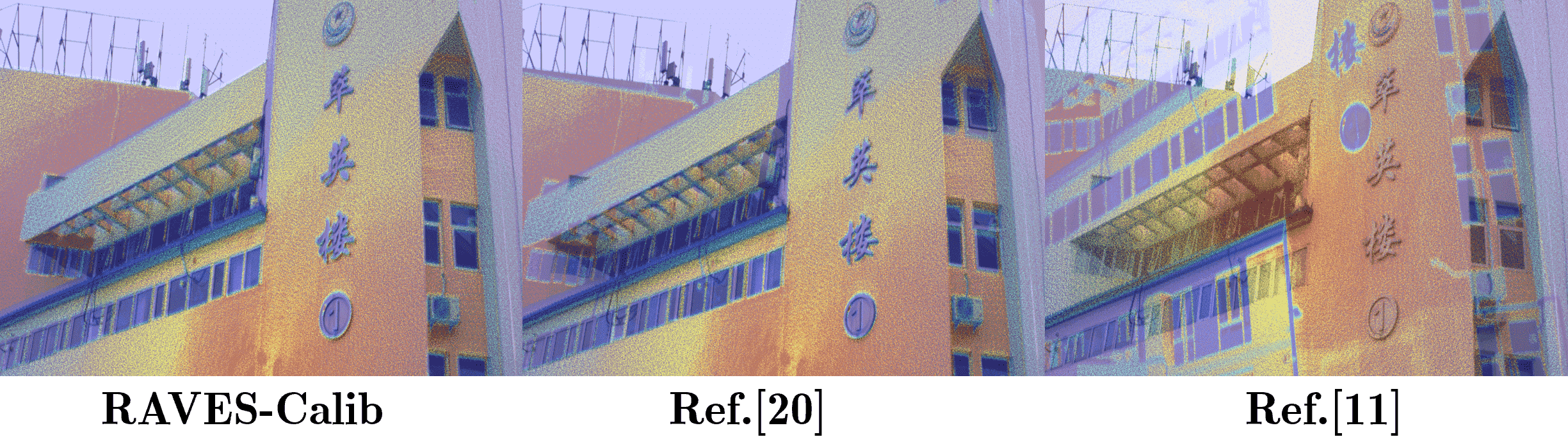}}
\caption{Comparison of initial value estimation results.}
\label{fig7}
\end{figure}

\subsection{Validation of the Initial Estimation Module}
\subsubsection{Robustness Verification}
We evaluated the robustness of the initial estimation method using six challenging data pairs collected in parking lots and offices. These challenges include: a large and completely unknown rotation angle between sensors exists, significant differences in FOV of LiDAR and camera, and semi-structured scenes. The first line of Fig.\ref{fig9} shows the experimental data. We used the methods in \cite{yuan2021pixel}, \cite{koide2023general}, and RAVES-Calib to perform initial estimation experiments on 6 sets of data containing four sensors, respectively. If the results of the initial estimate are sufficient to meet the needs of the refining part, we consider the initial estimate to be successful. As shown in Tab.\ref{tab2}, similar to \cite{koide2023general}, RAVES-Calib can complete initialization in most cases. However, the effect on the spinning LiDAR is slightly worse, because the overlap FOV between the sensors is small, and the reconstruction error of the spinning LiDAR is large. The initial value estimation method from \cite{yuan2021pixel} determines the initial value by searching for the pose with the most matching edge features within a limited pose range. However, this method is not well-suited for scenarios involving large rotation changes. So it could only complete the estimation for scenes $\left(4 \right)$,  $\left(5 \right)$, and was unable to handle the other four. We used our initial estimation result to project the LiDAR point cloud onto the camera image as shown in Fig.\ref{fig7}, qualitatively demonstrating the accuracy of the initial value. RAVES-Calib is not only able to adapt to large rotational variations but also improves its accuracy further due to the addition of line features. 


\subsubsection{Consistent Verification}
\begin{figure}[h]
	\begin{minipage}{0.48\linewidth}
		\vspace{3pt}
		\centerline{\includegraphics[width=\textwidth]{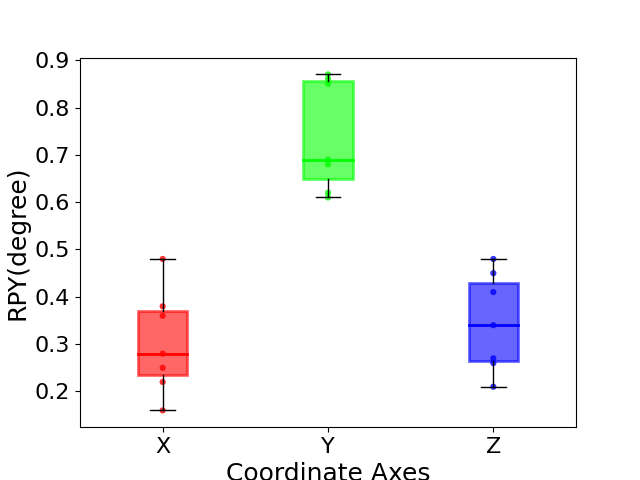}}
	\end{minipage}
	\begin{minipage}{0.48\linewidth}
		\vspace{3pt}
		\centerline{\includegraphics[width=\textwidth]{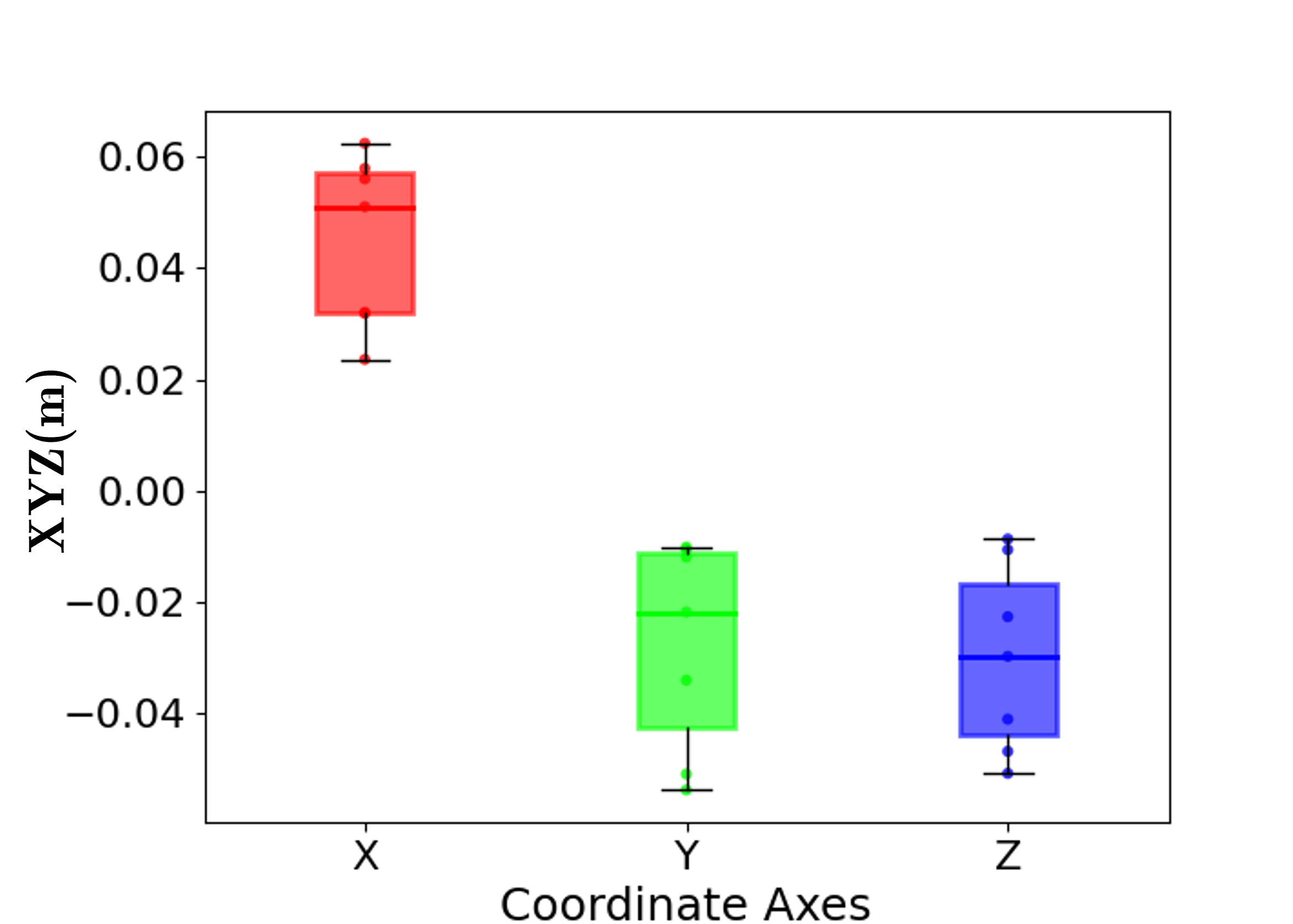}}
	\end{minipage}
	\caption{ Distribution of converged extrinsic values across all 10 scenarios. The horizontal axes depict the three coordinate directions (X, Y, Z), while the vertical axes indicate the error (rotation error for the left plot, translation error for the right plot).}
	\label{fig8}
\end{figure}

\begin{figure*}[htbp]
\includegraphics[width=0.985\linewidth]{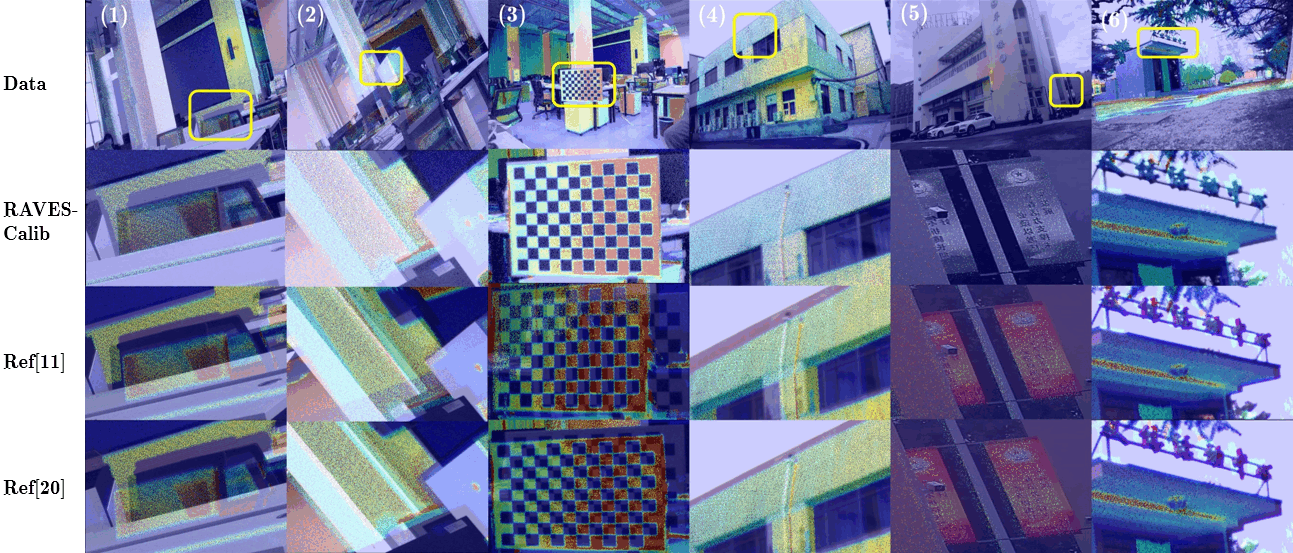}
\caption{Comparison of experimental results in various scenarios. The first rows show the Calibration scene. (1)-(3) are indoor scenes, (4)-(6) are outdoor scenes. (1)-(2) have a large rotation between the camera and the LiDAR. (6) is semi-structured scenes.}
\label{fig9}
\end{figure*}

The initial value should not only adapt to the specific scenario but should maintain consistency across different scenes. To verify consistency, we gather the statistics of initial estimation results of the SHARE camera and Livox AVIA conducted on 10 sets of data, 4 from the outdoors and the other 3 from the indoors. We tested each dataset 10 times. The statistical results are shown in Fig.\ref{fig8}. It can be seen that, across all scenarios, the estimated initial extrinsic parameters remain nearly consistent, reflecting the robustness and reliability of the initial value estimation process.

\subsection{Validation of the Optimization Module}
\subsubsection{Assessment Method}
We further evaluated the accuracy of RAVES-Calib by comparing it to the benchmark method from \cite{yuan2021pixel} and \cite{koide2023general}, SOTA targetless method known for achieving accuracy comparable to target-based methods. In practical deployment scenarios, the acquisition of groundtruth for extrinsic parameters between LiDAR and camera systems presents inherent challenges. The RAVES-Calib was evaluated through the evaluation method in \cite{cui2023aclc}, quantitative analysis of projection errors of checkerboard corner points between LiDAR intensity image and camera image. Checkerboard corner points in the image are used as pseudo groundtruth. The LiDAR point cloud containing the checkerboard is projected onto the image according to the estimated extrinsic parameters $E$ and camera intrinsic parameters $K$, and then the projection error is calculated. Due to the perspective projection of the camera, the distance between the checkerboard and the sensor will affect the scale of the projection point, therefore, we scale the reprojection by distance normalization to reduce this bias. The normalized reprojection errors (NRE) as described in Eq.(\ref{eq8}):
 \begin{equation}
    NRE=\sum_{p\in C^{\prime}_{3D}, c_{2d}\in C_{2D}} \frac{d\left( p \right)}{d_{max}}\left( \left| p-c_{2d} \right| \right)
    \label{eq8}  
 \end{equation}

\begin{equation}
C^{\prime}_{3D} = \frac{1}{z} \boldsymbol EK 
\left[ \begin{array}{c}
\boldsymbol C^{x}_{3D}\\
\boldsymbol C^{y}_{3D}\\
\boldsymbol C^{z}_{3D}
\end{array} 
\right ]
\label{eq9}  
\end{equation}

Where $C_{3D}$ and $C_{2D}$ is the 3D and 2D corners of checkerboard. $c_{2d}$ is the closest image pixel to the reprojected point $p$, and $d\left( p \right)$ is the distance of $p$ from LiDAR center. Take the average of all corner point errors as the final error. Fig.\ref{fig8} shows the checkerboard corner point detection results for point clouds and images, as well as the projection results.

\subsubsection{Accuracy Assessment}
We evaluated the accuracy of the RAVES-Calib using seven sets of data containing four sensors. The data scenarios include indoor, outdoor, structured, semi-structured, and unstructured scenarios. For data that resulted in a failed initial guess, we provide an initial estimate of the results from other data for evaluation of the fine registration algorithm. The seventh set of data is in an unstructured environment, and none of the three methods can complete the calibration. RAVES-Calib, \cite{yuan2021pixel} and \cite{koide2023general} worked well for the initial six datasets. The initial six sets of data scenarios and the details of calibration results by using three methods are shown in Fig.\ref{fig9}. It shows that in most scenarios, three methods achieve high-accuracy calibration, particularly in environments with distinct structural features and uniformly distributed features. However, thanks to the optimal feature distribution-based optimization, RAVES-Calib outperformed the benchmark in handling local details.

The NER quantitative evaluation was used to further validate the optimization module, and the result as shown in Tab.\ref{tab3} and Tab.\ref{tab4}. RAVES-Calib achieved the ideal calibration error on average for different sensor combinations: solid-state LiDAR and full frame camera 4.325 Pixel, solid-state LiDAR and pinhole camera 5.347 Pixel, spinning LiDAR and full frame Camera 5.984 Pixel, spinning LiDAR and pinhole camera 6.205 Pixel. The calibration results of each sensor are superior to those of the other methods. As can be seen in Tab.\ref{tab4}, \cite{yuan2021pixel} and \cite{koide2023general} results of the third set of data fluctuate greatly, but the RAVES-Calib results are stable because it can use multiple types of features (chessboard corners as point features). Since RAVES-Calib and \cite{yuan2021pixel} are sensitive to edge features of the environment, there are fluctuations in the semi-structured environment(Fig.\ref{fig9}$\left(6 \right)$). However, our method combined with point features still maintains a high accuracy.

 \begin{figure}[htbp]
\centerline{\includegraphics[width=1\linewidth]{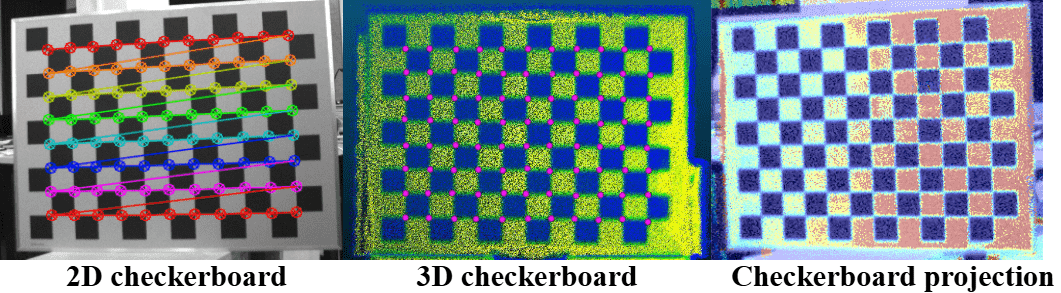}}
\caption{Visualization of the estimated corners from the checkerboard.}
\label{fig10}
\end{figure}

\begin{table}[bt]
\caption{QUANTITATIVE EVALUATION OF THE CALIBRATION METHODS WITH SOLID-STATE LIDAR AND
CAMERA. USE THE AVERAGE NORMALIZED REPROJECTION ERROR(PIXELS) AS THE RESULT.}
\begin{center}
\resizebox{8.6cm}{!}{
\begin{tabular}{|c|c|c|c|c|c|c|}
\hline
\textbf{Data} 
&\multicolumn{3}{|c|}{\textbf{\makecell{Solid-state LiDAR\\Full Frame Camera}}} &\multicolumn{3}{|c|}{\textbf{\makecell{Solid-state LiDAR\\Pinhole Camera}}}\\
 \cline{2-4} \cline{5-7}
 \text{}  & \textbf{\text{RAVES}}& \textbf{\text{Ref\cite{yuan2021pixel}}} &\textbf{\textbf{Ref\cite{koide2023general}}}&\textbf{\text{RAVES}}&\textbf{\textbf{Ref\cite{yuan2021pixel}}}&\textbf{\textbf{Ref\cite{koide2023general}}}\\
\hline
01&  \textbf{3.22} & 9.93  & 8.26 & \textbf{4.46} & 11.64 & 9.25  \\
02&  \textbf{3.42} & 12.52 & 5.52 & \textbf{3.64} & 14.25 & 4.83  \\
03&  \textbf{2.24} & 22.93 & 20.82& \textbf{3.42} & 24.22 & 24.33\\
04&  \textbf{5.53} & 15.96 & 9.09 & \textbf{7.31} & 18.36 & 12.35 \\
05&  \textbf{4.22} & 8.13  & 10.06& \textbf{5.84} & 9.28  & 9.35 \\
06&  \textbf{7.32} & 13.54 & 17.83& \textbf{7.41} & 15.26 &22.09 \\
07&$\times$ &$\times$ & $\times$& $\times$&  $\times$& $\times$\\
Avg.&\textbf{4.325} & 13.835& 11.93& \textbf{5.347}& 15.501 &13.7\\
\hline

\end{tabular}
}

\label{tab3}
\end{center}
\end{table}

\begin{table}[bt]
\caption{QUANTITATIVE EVALUATION OF THE CALIBRATION METHODS WITH SPINNING LIDAR AND
CAMERA.}
\begin{center}
\resizebox{8.6cm}{!}{
\begin{tabular}{|c|c|c|c|c|c|c|c|}
\hline
\textbf{Data}  
&\multicolumn{3}{|c|}{\textbf{\makecell{Spinning LiDAR\\Full Frame Camera}}} &\multicolumn{3}{|c|}{\textbf{\makecell{Spinning LiDAR\\Pinhole Camera}}}\\
 \cline{2-4} \cline{5-7}
 \text{}  & \textbf{\textbf{RAVES}}& \textbf{\textbf{Ref\cite{koide2023general}}} &\textbf{\textbf{Ref\cite{yuan2021pixel}}}&\textbf{\text{RAVES}}&\textbf{\textbf{Ref\cite{koide2023general}}}&\textbf{\textbf{Ref\cite{yuan2021pixel}}}\\
\hline
01& \textbf{5.29}  & 10.93 & 8.03 & \textbf{4.45}  & 5.98 & 4.72  \\
02& \textbf{5.23}  & 12.45 & 9.23 & \textbf{4.98}  & 7.23 & 5.29  \\
03& \textbf{3.56}  & 22.59 & 18.25& \textbf{4.23}  & 23.98 & 21.43\\
04& \textbf{4.36}  & 11.21 & 8.98 & \textbf{7.13}  & 12.42 & 8.79 \\
05& \textbf{7.34}  & 9.29  & 11.84& \textbf{5.05}  & 14.28 & 9.12 \\
06& \textbf{10.12} & 12.22 & 15.21& \textbf{11.39} & 12.12 & 12.23 \\
07&$\times$ &$\times$ & $\times$& $\times$&  $\times$& $\times$\\
Avg.&\textbf{5.984} & 13.115& 11.923& \textbf{6.205}&  12.668& 10.264\\
\hline

\end{tabular}
}

\label{tab4}
\end{center}
\end{table}

\subsection{Ablation Experiment}
\begin{figure}[tb]
\centerline{\includegraphics[width=\linewidth]{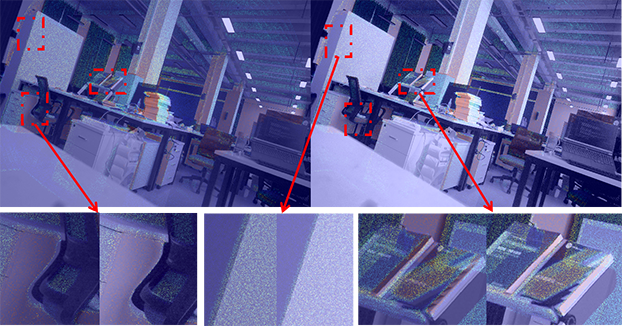}}
\caption{Fused image generated by projecting LiDAR points onto the camera image based on the final extrinsic parameters obtained through calibration. The left image shows results without the feature distribution analysis, while the right shows results with it. The lower images highlight detailed comparisons.}
\label{fig11}
\end{figure}

To assess the effectiveness of the proposed feature distribution-based optimization scheme, we conducted an ablation study. Using the same dataset, we performed two tests: one without the feature distribution analysis module, utilizing all features uniformly, and another incorporating the adaptive weighting scheme to suppress the influence of inferior features on the calibration optimization. The results of this comparison are shown in Fig.\ref{fig11}. Using NER for quantitative analysis, the results are shown in Tab.\ref{tab5}.

In both cases, satisfactory calibration results were achieved. However, closer examination reveals that the method incorporating feature distribution analysis provides better alignment of local object edges and improves the overall calibration accuracy by about one pixel.

\begin{table}[bt]
\caption{QUANTITATIVE EVALUATION OF ABLATION EXPERIMENT.}
\begin{center}
\resizebox{8.6cm}{!}{
\begin{tabular}{|c|c|c|c|c|}
\hline
\textbf{Contribute}&\textbf{\text{\makecell{Solid-state\\Full Frame}}} &\textbf{\text{\makecell{Solid-state \\Pinhole Camera }}} &\textbf{\text{\makecell{Spinning \\Full Frame }}} &\textbf{\text{\makecell{Spinning \\Pinhole Camera }}} \\

\hline
$\times$& 4.35 & 5.74 & 5.38 & 6.69 \\

$\checkmark$& \textbf{3.24} & \textbf{4.34} & \textbf{5.23} & \textbf{5.98}\\
\hline

\end{tabular}
}

\label{tab5}
\end{center}
\end{table}

\section{CONCLUSIONS}
This paper presents a targetless extrinsic calibration toolkit, RAVES-Calib, that integrates stable point and line features extracted and matched by Gluestick. To enhance the stability and accuracy of the calibration, the effect of features distribution on calibration estimation is comprehensively analyzed, guiding the optimization convergence. The proposed toolkit is validated in both indoor and outdoor environments, achieving accuracy comparable to target-based methods.

\addtolength{\textheight}{-7cm}   




\bibliographystyle{IEEEtran}
\bibliography{IEEEabrv,ref}

\begin{thebibliography}{10}
\providecommand{\url}[1]{#1}
\csname url@samestyle\endcsname
\providecommand{\newblock}{\relax}
\providecommand{\bibinfo}[2]{#2}
\providecommand{\BIBentrySTDinterwordspacing}{\spaceskip=0pt\relax}
\providecommand{\BIBentryALTinterwordstretchfactor}{4}
\providecommand{\BIBentryALTinterwordspacing}{\spaceskip=\fontdimen2\font plus
\BIBentryALTinterwordstretchfactor\fontdimen3\font minus \fontdimen4\font\relax}
\providecommand{\BIBforeignlanguage}[2]{{%
\expandafter\ifx\csname l@#1\endcsname\relax
\typeout{** WARNING: IEEEtran.bst: No hyphenation pattern has been}%
\typeout{** loaded for the language `#1'. Using the pattern for}%
\typeout{** the default language instead.}%
\else
\language=\csname l@#1\endcsname
\fi
#2}}
\providecommand{\BIBdecl}{\relax}
\BIBdecl

\bibitem{puente2013review}
I.~Puente, H.~Gonz{\'a}lez-Jorge, J.~Mart{\'\i}nez-S{\'a}nchez, and P.~Arias, ``Review of mobile mapping and surveying technologies,'' \emph{Measurement}, vol.~46, no.~7, pp. 2127--2145, 2013.

\bibitem{lin2024r}
J.~Lin and F.~Zhang, ``R$^3$live$++$: A robust, real-time, radiance reconstruction package with a tightly-coupled lidar-inertial-visual state estimator,'' \emph{IEEE Transactions on Pattern Analysis and Machine Intelligence}, 2024.

\bibitem{4432828}
A.~Discant, A.~Rogozan, C.~Rusu, and A.~Bensrhair, ``Sensors for obstacle detection - a survey,'' in \emph{2007 30th International Spring Seminar on Electronics Technology (ISSE)}, 2007, pp. 100--105.

\bibitem{10089400}
A.~Pandharipande, C.-H. Cheng, J.~Dauwels, S.~Z. Gurbuz, J.~Ibanez-Guzman, G.~Li, A.~Piazzoni, P.~Wang, and A.~Santra, ``Sensing and machine learning for automotive perception: A review,'' \emph{IEEE Sensors Journal}, vol.~23, no.~11, pp. 11\,097--11\,115, 2023.

\bibitem{yang2015automatic}
B.~Yang and C.~Chen, ``Automatic registration of uav-borne sequent images and lidar data,'' \emph{ISPRS Journal of Photogrammetry and Remote Sensing}, vol. 101, pp. 262--274, 2015.

\bibitem{li2019nrli}
J.~Li, B.~Yang, C.~Chen, and A.~Habib, ``Nrli-uav: Non-rigid registration of sequential raw laser scans and images for low-cost uav lidar point cloud quality improvement,'' \emph{ISPRS Journal of Photogrammetry and Remote Sensing}, vol. 158, pp. 123--145, 2019.

\bibitem{huang2024novel}
Z.~Huang, X.~Zhang, A.~Garcia, and X.~Huang, ``A novel, efficient and accurate method for lidar camera calibration,'' in \emph{2024 IEEE International Conference on Robotics and Automation (ICRA)}.\hskip 1em plus 0.5em minus 0.4em\relax IEEE, 2024, pp. 14\,513--14\,519.

\bibitem{domhof2021joint}
J.~Domhof, J.~F. Kooij, and D.~M. Gavrila, ``A joint extrinsic calibration tool for radar, camera and lidar,'' \emph{IEEE Transactions on Intelligent Vehicles}, vol.~6, no.~3, pp. 571--582, 2021.

\bibitem{yi2022extrinsic}
H.~Yi, B.~Liu, B.~Zhao, and E.~Liu, ``Extrinsic calibration for lidar--camera systems using direct 3d--2d correspondences,'' \emph{Remote Sensing}, vol.~14, no.~23, p. 6082, 2022.

\bibitem{liao2018extrinsic}
Q.~Liao, Z.~Chen, Y.~Liu, Z.~Wang, and M.~Liu, ``Extrinsic calibration of lidar and camera with polygon,'' in \emph{2018 IEEE International Conference on Robotics and Biomimetics (ROBIO)}.\hskip 1em plus 0.5em minus 0.4em\relax IEEE, 2018, pp. 200--205.

\bibitem{yuan2021pixel}
C.~Yuan, X.~Liu, X.~Hong, and F.~Zhang, ``Pixel-level extrinsic self calibration of high resolution lidar and camera in targetless environments,'' \emph{IEEE Robotics and Automation Letters}, vol.~6, no.~4, pp. 7517--7524, 2021.

\bibitem{park2014calibration}
Y.~Park, S.~Yun, C.~S. Won, K.~Cho, K.~Um, and S.~Sim, ``Calibration between color camera and 3d lidar instruments with a polygonal planar board,'' \emph{Sensors}, vol.~14, no.~3, pp. 5333--5353, 2014.

\bibitem{kwak2011extrinsic}
K.~Kwak, D.~F. Huber, H.~Badino, and T.~Kanade, ``Extrinsic calibration of a single line scanning lidar and a camera,'' in \emph{2011 IEEE/RSJ International Conference on Intelligent Robots and Systems}.\hskip 1em plus 0.5em minus 0.4em\relax IEEE, 2011, pp. 3283--3289.

\bibitem{6232233}
L.~Zhou and Z.~Deng, ``Extrinsic calibration of a camera and a lidar based on decoupling the rotation from the translation,'' in \emph{2012 IEEE Intelligent Vehicles Symposium}, 2012, pp. 642--648.

\bibitem{mirzaei20123d}
F.~M. Mirzaei, D.~G. Kottas, and S.~I. Roumeliotis, ``3d lidar--camera intrinsic and extrinsic calibration: Identifiability and analytical least-squares-based initialization,'' \emph{The International Journal of Robotics Research}, vol.~31, no.~4, pp. 452--467, 2012.

\bibitem{cui2023aclc}
J.~Cui, J.~Niu, Y.~He, D.~Liu, and Z.~Ouyang, ``Aclc: Automatic calibration for nonrepetitive scanning lidar-camera system based on point cloud noise optimization,'' \emph{IEEE Transactions on Instrumentation and Measurement}, vol.~73, pp. 1--14, 2023.

\bibitem{wu2022lidar}
Q.~Wu, J.~Zhang, J.~Sheng, C.~Wu, and H.~Yuan, ``Lidar-camera system automatic extrinsic calibration in rail transit,'' in \emph{2022 IEEE 25th International Conference on Intelligent Transportation Systems (ITSC)}.\hskip 1em plus 0.5em minus 0.4em\relax IEEE, 2022, pp. 3380--3385.

\bibitem{bai2020lidar}
Z.~Bai, G.~Jiang, and A.~Xu, ``Lidar-camera calibration using line correspondences,'' \emph{Sensors}, vol.~20, no.~21, p. 6319, 2020.

\bibitem{munoz2020targetless}
M.~A. Munoz-Banon, F.~A. Candelas, and F.~Torres, ``Targetless camera-lidar calibration in unstructured environments,'' \emph{IEEE Access}, vol.~8, pp. 143\,692--143\,705, 2020.

\bibitem{koide2023general}
K.~Koide, S.~Oishi, M.~Yokozuka, and A.~Banno, ``General, single-shot, target-less, and automatic lidar-camera extrinsic calibration toolbox,'' in \emph{2023 IEEE International Conference on Robotics and Automation (ICRA)}.\hskip 1em plus 0.5em minus 0.4em\relax IEEE, 2023, pp. 11\,301--11\,307.

\bibitem{mastin2009automatic}
A.~Mastin, J.~Kepner, and J.~Fisher, ``Automatic registration of lidar and optical images of urban scenes,'' in \emph{2009 IEEE conference on computer vision and pattern recognition}.\hskip 1em plus 0.5em minus 0.4em\relax IEEE, 2009, pp. 2639--2646.

\bibitem{huang2024onlinetargetfreelidarcameraextrinsiccalibration}
\BIBentryALTinterwordspacing
Z.~Huang, Y.~Zhang, Q.~Chen, and R.~Fan, ``Online,target-free lidar-camera extrinsic calibration via cross-modal mask matching,'' 2024. [Online]. Available: \url{https://arxiv.org/abs/2404.18083}
\BIBentrySTDinterwordspacing

\bibitem{li2023automatic}
X.~Li, Y.~Xiao, B.~Wang, H.~Ren, Y.~Zhang, and J.~Ji, ``Automatic targetless lidar--camera calibration: a survey,'' \emph{Artificial Intelligence Review}, vol.~56, no.~9, pp. 9949--9987, 2023.

\bibitem{SEO2004733}
\BIBentryALTinterwordspacing
J.-K. Seo, H.-K. Hong, C.-W. Jho, and M.-H. Choi, ``Two quantitative measures of inlier distributions for precise fundamental matrix estimation,'' \emph{Pattern Recognition Letters}, vol.~25, no.~6, pp. 733--741, 2004. [Online]. Available: \url{https://www.sciencedirect.com/science/article/pii/S0167865504000273}
\BIBentrySTDinterwordspacing

\bibitem{X-ICP}
T.~Tuna, J.~Nubert, Y.~Nava, S.~Khattak, and M.~Hutter, ``X-icp: Localizability-aware lidar registration for robust localization in extreme environments,'' \emph{IEEE Transactions on Robotics}, vol.~40, pp. 452--471, 2024.

\bibitem{pautrat2023gluestick}
R.~Pautrat, I.~Su{\'a}rez, Y.~Yu, M.~Pollefeys, and V.~Larsson, ``Gluestick: Robust image matching by sticking points and lines together,'' in \emph{Proceedings of the IEEE/CVF International Conference on Computer Vision}, 2023, pp. 9706--9716.

\bibitem{1240258}
N.~Gelfand, L.~Ikemoto, S.~Rusinkiewicz, and M.~Levoy, ``Geometrically stable sampling for the icp algorithm,'' in \emph{Fourth International Conference on 3-D Digital Imaging and Modeling, 2003. 3DIM 2003. Proceedings.}, 2003, pp. 260--267.

\bibitem{10.1115/1.4031335}
\BIBentryALTinterwordspacing
T.-H. Kwok and K.~Tang, ``{Improvements to the Iterative Closest Point Algorithm for Shape Registration in Manufacturing},'' \emph{Journal of Manufacturing Science and Engineering}, vol. 138, no.~1, p. 011014, 09 2015. [Online]. Available: \url{https://doi.org/10.1115/1.4031335}
\BIBentrySTDinterwordspacing

\bibitem{8301593}
T.-H. Kwok, ``Dnss: Dual-normal-space sampling for 3-d icp registration,'' \emph{IEEE Transactions on Automation Science and Engineering}, vol.~16, no.~1, pp. 241--252, 2019.

\end{thebibliography}

\end{document}